# Robust Motion Control for Mobile Manipulator Using Resolved Acceleration and Proportional-Integral Active Force Control

**Musa Mailah, Endra Pitowarno & Hishamuddin Jamaluddin**
Department of Applied Mechanics, Faculty of Mechanical Engineering
Universiti Teknologi Malaysia (UTM), 81310 Skudai – JB, Malaysia
musa@fkm.utm.my, epit@mech.fkm.utm.my/epit@eepis-its.edu, hishamj@fkm.utm.my

*Abstract: A resolved acceleration control (RAC) and proportional-integral active force control (PIAFC) is proposed as an approach for the robust motion control of a mobile manipulator (MM) comprising a differentially driven wheeled mobile platform with a two-link planar arm mounted on top of the platform. The study emphasizes on the integrated kinematic and dynamic control strategy in which the RAC is used to manipulate the kinematic component while the PIAFC is implemented to compensate the dynamic effects including the bounded known/unknown disturbances and uncertainties. The effectivenss and robustness of the proposed scheme are investigated through a rigorous simulation study and later complemented with experimental results obtained through a number of experiments performed on a fully developed working prototype in a laboratory environment. A number of disturbances in the form of vibratory and impact forces are deliberately introduced into the system to evaluate the system performances. The investigation clearly demonstrates the extreme robustness feature of the proposed control scheme compared to other systems considered in the study.*

## 1. Introduction

A mobile manipulator is basically a robotic arm mounted on a moving base and can be used to perform variety of tasks that are mostly related to materials handling application. Various aspects of the mobile mobile manipulator have been studied particularly on the effective motion control of the system. The prerequisites of implementing a good overall control of the system often involves the study of kinematics and dynamics of the mobile manipulator.

There are a number of extensive works that can be found in literature related to the kinematic of the mobile manipulator (Perrier *et al.*, 1998, Bayle *et al.*, 2001, Sugar and Kumar, 2002, and Tanner, 2003). The kinematic analysis is particularly useful to describe the robot's workspace and motion path planning tasks including obstacles avoidance, collision free moving capability and maneuverability.

Perrier *et al.* (1998) implemented homogenous matrices and dual quarternion to represent the redundancies in the kinematic problems. They considered the global motion of mobile manipulator from point to point and computed a path that takes into account different constraints; the nonholonomic and holonomic. Their works were successfully used and applied to resolve the redundancy of the kinematics problem in joints limitation, velocity and radius steering limitations. Bayle *et al.* (2001) focused the investigation on the manipulability of a particular class of mobile manipulator in local kinematic analysis. They showed how the notion of manipulability could be extended to represent the operational methods in a configuration of the system. One of the advantages of their works is that it can be used to reconfigure the installed robot arms position on the top of the platform in order to maximize the workspace operations. Sugar and Kumar (2002) developed a kinematic-based control of multiple mobile manipulators. Their analysis was based on tasks that require grasping, manipulation and transporting large and possibly flexible objects without special purpose fixtures. The kinematic problem was solved using the compliant arms analysis and it was successfully demonstrated for cooperation of two and three mobile manipulators. Tanner (2003) proposed a new methodology to motion planning for multiple mobile manipulators cooperation that are applicable to articulated, non point nonholonomic robots with guaranteed collision avoidance and convergence properties. He implemented a potential field technique using *diffeomorphic* transformations and the resulting point-world topology. The approach was applied to multiple mobile manipulators in handling deformable material in an obstacle environment and it showed successfully through simulation. The main feature of the method is the application of dipolar potential field technique to guarantee the robot will approach the destination asymptotically and it will follow the path that automatically stabilizes its orientation. This method incorporated an inverse Lyapunov function to stabilize (and converge) the robots navigation. The works are very useful for future developments in coordinated kinematics controls, but unfortunately the robustness issue was not specifically addressed.

In actual implementation pertaining to robot's motion, it is also typical to address the control problem involving the robot's dynamic. At this juncture, designing appropriate controller can lead to significant improvement in performance (Yamamoto and Yun, 1996). Combining both the extensive kinematic and dynamic aspects for an ideal motion control of any dynamical system still remains a complex and challenging problem. In recent years, several researchers have contributed to solving this problem using a number of methods.

Colbaugh (1998) addressed the problem of stabilizing mobile manipulators in the presence of uncertainties regarding the system dynamic model. He proposed an effective solution by combining homogenous system theory and adaptive control theory that was theoretically proven. Mohri *et al.* (2001) presented a trajectory planning method using an optimal control theory. They derived the robot's dynamics by considering it as a combined system of mobile platform and manipulator. They then applied a sub-optimal trajectory planning using an iterative algorithm based on gradient functions synthesized in the hierarchical manner to formulate the trajectory-planning problem. They reported the effectiveness of the scheme by simulation. Lin and Goldenberg (2001) proposed a class of neural network control of mobile manipulator that is subjected to kinematics constraint. The study assumed that the robot's dynamic is completely unknown and it would be identified using a neural network estimator. For the trajectory tracking control, a class of feedback control was employed with its stability tested using a Lyapunov function theory. A procedure to estimate the dynamics was used by first redefining the robots dynamics as an error dynamics based on a set of carefully chosen Lyapunov sub functions through a joint-space tracking. Next, a neural network (NN) online estimator was constructed and a new-NN learning law was obtained from which a new NN control could be then derived. A simulation study proved the effectiveness of the proposed scheme.

In this paper, a simplified coordinate $(x, y)$ and heading angle $(\varphi)$ of a nonholonomic mobile manipulator motion control using resolved acceleration control (RAC) combined with proportional plus integral active force control (PIAFC) is proposed. The RAC part that was attributed to Luh *et al.* (1980) is a simple but yet powerful acceleration mode control method that could improve the performance of the existing conventional servo control as reported in a number of studies (Kircanski and Kircanski, 1998, Campa, *et al.*, 2001). It is still considered by many as one of the best control options due to its simplicity in real-time implementation and was developed as the integrated simplified mobile robot with coordinate and heading angle $(x_v, y_v, \varphi)$ control and the *XY Cartesian* planar manipulator's tip position coordinate $(x_m, y_m)$ control. By using this RAC-based $x$ and $y$ control, the proposed control scheme would have a more flexible position, speed and acceleration control. This flexibility is gained by the use of simultaneous reference input position, velocity and acceleration parameters.

To tackle the robot's dynamic problem particularly those involving disturbances and uncertainties, the proposed PIAFC scheme is implemented. The original work of AFC was attributed to Hewit and Burdess (1981). The overall control scheme is to be known as RAC-PIAFC and through this scheme, a simulation and experimental study on the robot model considering both the kinematic and dynamic effects was rigorously performed.

The paper firstly deals with the modelling of the MM followed by the proposed controller design procedure that contains the RAC, AFC and PIAFC elements. Then, the simulations performed are discussed to evaluate the control performances. Lastly, a basic practical experimentation and evaluation of a MM prototype endowed with the proposed control method is presented to demonstrate the feasibility of the scheme.

## 2. Modelling of a Mobile Manipulator

Consider a mobile manipulator depicted in Fig. 1. It consists of a wheeled mobile platform and a two-link manipulator mounted on top of the platform.

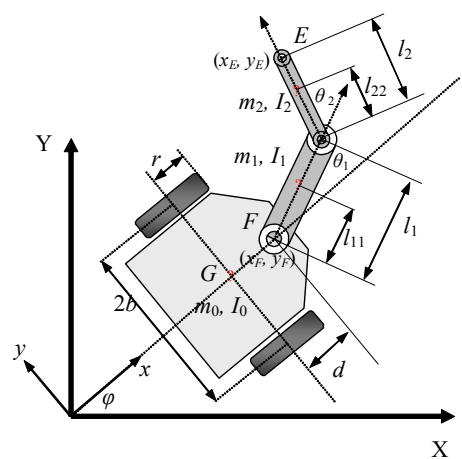

Fig. 1. A mobile manipulator

The platform moves by driving the two independent wheels. The manipulator is constructed as a two-link planar arm with servo motors attached at the joints. *XY* and *xy* are the world and robot coordinate systems respectively, $\varphi$ is the robot's heading angle, $b$ is the width of the robot, $r$ is the radius of the wheel, and $d$ is the distance between point G and F. The mass and inertia of the platform are denoted as $m_0$ and $I_0$ respectively. For the manipulator, $l_1$ is the length of link-1, $l_{11}$ is the distance between F to the center of the mass of link-1, $l_2$ is the length of link-2, $l_{22}$ is the distance between joint-2 to the centre of the mass of link-2. $m_1$, $m_2$ and $I_1$, $I_2$ represent the masses and inertias of link-1 and link-2 respectively. The coordinate of the tip position is denoted by $(x_E, y_E)$. It is assumed that the velocity at which this system moves is relatively slow and thus the two driven wheels do not slip sideways. The velocity of the platform at the centre of mass, $v_G$, is then perpendicular to the wheel axis. This expresses $x$ and $y$ components in a nonholonomic manner described by the following equation,

$$\dot{x}_G \sin\varphi - \dot{y}_G \cos\varphi = 0 \tag{1}$$

For point F, the constraint can be written as

$$\dot{x}_F \sin\varphi - \dot{y}_F \cos\varphi + \dot{\varphi}\, d = 0 \tag{2}$$

The kinematic equation of the platform can be expressed as

$$\begin{pmatrix} \dot{x}_F \\ \dot{y}_F \\ \dot{\varphi}_F \end{pmatrix} = \begin{pmatrix} \dfrac{r}{2}\cos\varphi + \dfrac{d\cdot r}{2b}\sin\varphi & \dfrac{r}{2}\cos\varphi - \dfrac{d\cdot r}{2b}\sin\varphi \\ \dfrac{r}{2}\sin\varphi - \dfrac{d\cdot r}{2b}\cos\varphi & \dfrac{r}{2}\sin\varphi + \dfrac{d\cdot r}{2b}\cos\varphi \\ -\dfrac{r}{2b} & \dfrac{r}{2b} \end{pmatrix} \begin{pmatrix} \dot{\theta}_L \\ \dot{\theta}_R \end{pmatrix} \tag{3}$$

The rotation matrix is explicitly given by

$$\begin{pmatrix} \dot{x}_F \\ \dot{y}_F \end{pmatrix} = \begin{pmatrix} \cos\varphi & -\sin\varphi \\ \sin\varphi & \cos\varphi \end{pmatrix} \begin{pmatrix} \dfrac{r}{2} & \dfrac{r}{2} \\ -\dfrac{d\cdot r}{2b} & \dfrac{d\cdot r}{2b} \end{pmatrix} \begin{pmatrix} \dot{\theta}_L \\ \dot{\theta}_R \end{pmatrix} \tag{4}$$

For the manipulator mounted on board the platform at point $F$, its forward kinematic can be described as

$$\begin{pmatrix} \dot{x}_E \\ \dot{y}_E \end{pmatrix} = \begin{pmatrix} \dot{x}_F \\ \dot{y}_F \end{pmatrix} + \begin{pmatrix} \cos\varphi & -\sin\varphi \\ \sin\varphi & \cos\varphi \end{pmatrix} \begin{pmatrix} J_{11} & J_{12} \\ J_{21} & J_{22} \end{pmatrix} \begin{pmatrix} \dot{\theta}_1 + \dot{\varphi} \\ \dot{\theta}_2 \end{pmatrix} \tag{5}$$

where

$$J_{11} = -l_1 \sin\theta_1 - l_2 \sin(\theta_1 + \theta_2)$$
$$J_{12} = -l_2 \sin(\theta_1 + \theta_2)$$
$$J_{21} = l_1 \cos\theta_1 + l_2 \cos(\theta_1 + \theta_2)$$
$$J_{22} = l_2 \cos(\theta_1 + \theta_2)$$

Eq. (5) indicates that the kinematic control of the two sub-systems (platform and manipulator) can be partially solved. It is sometimes very useful to analyze the redundancy of the system when the robot arm is out of reach beyond its workspace. If $(x_F, y_F)$ is assumed to be

in a fixed position (platform is not moving and hence $\dot{\varphi}$ = 0), Eq. (5) can be expressed as

$$\begin{pmatrix} \dot{x}_T \\ \dot{y}_T \end{pmatrix} = \begin{pmatrix} J_{11} & J_{12} \\ J_{21} & J_{22} \end{pmatrix} \begin{pmatrix} \dot{\theta}_1 \\ \dot{\theta}_2 \end{pmatrix} \tag{6}$$

where $(x_T, y_T)$ is the tip position coordinate relative to the workspace of the manipulator. Then its inverse kinematic is

$$\begin{pmatrix} \dot{\theta}_1 \\ \dot{\theta}_2 \end{pmatrix} = \begin{pmatrix} J_{11} & J_{12} \\ J_{21} & J_{22} \end{pmatrix}^{-1} \begin{pmatrix} \dot{x}_T \\ \dot{y}_T \end{pmatrix} \tag{7}$$

From Eqs. (4) to (7) and letting $q = [\, x_F, y_F, x_E, y_E \,]^T$ as the input reference coordinate, the total kinematic equation of the mobile manipulator is

$$\begin{bmatrix} \dot{x}_E \\ \dot{y}_E \\ \dot{x}_F \\ \dot{y}_F \end{bmatrix} = \begin{bmatrix} \cos\varphi & -\sin\varphi & 0 & 0 \\ \sin\varphi & \cos\varphi & 0 & 0 \\ 0 & 0 & \cos\varphi & -\sin\varphi \\ 0 & 0 & \sin\varphi & \cos\varphi \end{bmatrix}$$
$$\begin{bmatrix} \dfrac{r}{2} - J_{11}\dfrac{r}{b} & \dfrac{r}{2} + J_{11}\dfrac{r}{b} & J_{11} & J_{12} \\ -(d + J_{21})\dfrac{r}{b} & (d + J_{21})\dfrac{r}{b} & J_{21} & J_{22} \\ \dfrac{r}{2} & \dfrac{r}{2} & 0 & 0 \\ -d\dfrac{r}{b} & d\dfrac{r}{b} & 0 & 0 \end{bmatrix} \begin{bmatrix} \dot{\theta}_L \\ \dot{\theta}_R \\ \dot{\theta}_1 \\ \dot{\theta}_2 \end{bmatrix} \tag{8}$$

where $\dot{\theta}_L$ and $\dot{\theta}_R$ are the angular velocities of the left and right wheels respectively, $\dot{\theta}_1$ and $\dot{\theta}_2$ are the angular velocities of the joints at link-1 and link-2 respectively. A mobile manipulator dynamic equation can be obtained using the Lagrangian approach (Yamamoto and Yun, 1996, Lin and Goldenberg, 2001) in the form,

$$\mathbf{M}(q)\ddot{q} + \mathbf{C}(q,\dot{q})\dot{q} + \mathbf{F}(q,\dot{q}) + \mathbf{A}^T(q)\lambda + \tau_d = \mathbf{B}(q)\tau \tag{9}$$

where

$q \in \Re^p$     is a generalized coordinate

$\mathbf{M}(q) \in \Re^{p \times p}$    is a symmetric and positive definite inertia matrix

$\mathbf{C}(q,\dot{q}) \in \Re^{p \times p}$   is the centripetal and Corioli's matrix

$\mathbf{F}(q,\dot{q}) \in \Re^p$    is the friction and gravitational vector

$\mathbf{A}(q) \in \Re^{r \times p}$     is a constraint matrix

$\lambda \in \Re^r$      is the Lagrange multiplier which denotes the vector of constraint forces

$\tau_d \in \Re^p$      is the bounded unknown disturbances including unstructured dynamics

$\mathbf{B}(q) \in \Re^{p \times (p-r)}$ is the input transformation matrix

$\tau \in \Re^{p-r}$      is the torques input vector

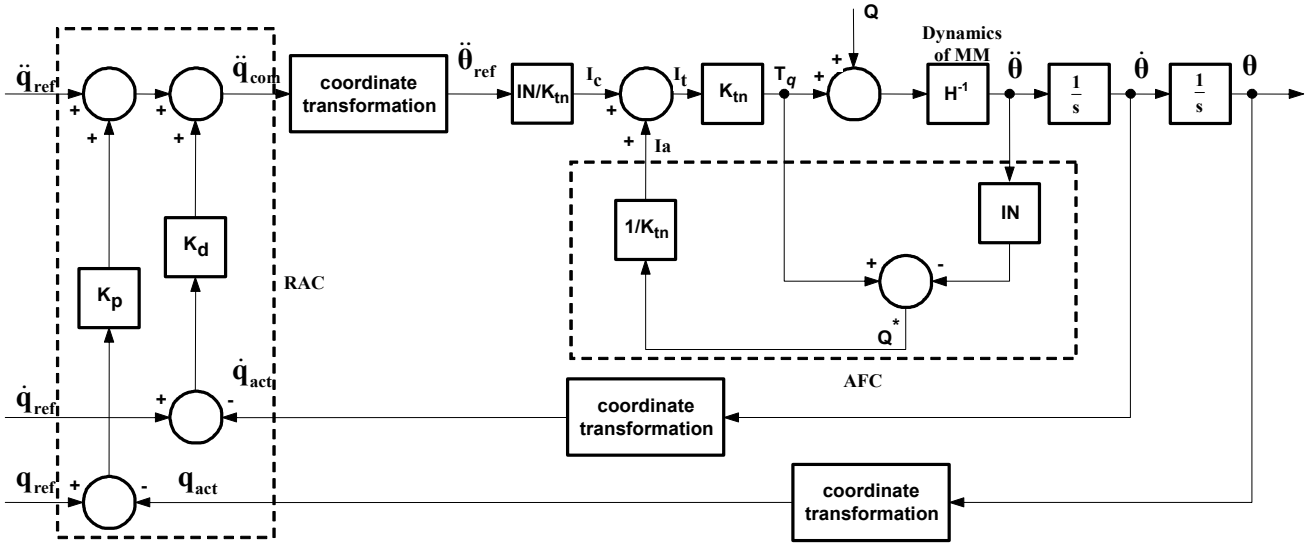

Fig. 2. The proposed controller

For the mobile manipulator considered in the study, the components of friction and gravitational vectors are neglected. Considering the well-known dynamic model of the manipulator, Eq. (9) can be expressed in terms of the dynamic interaction and coupling as follows,

$$\mathbf{M}_r(q_r)\ddot{q}_r + \mathbf{C}_{r1}(q_r,\dot{q}_r) + \mathbf{C}_{r2}(q_r,\dot{q}_r,\dot{q}_v) = \boldsymbol{\tau}_r - \mathbf{R}_v(q_r q_v)\ddot{q}_v \quad (10)$$

$$\mathbf{M}_{v1}(q_v)\ddot{q}_v + \mathbf{C}_{v1}(q_v,\dot{q}_v) + \mathbf{C}_{v2}(q_r,q_v,\dot{q}_r,\dot{q}_v) = \\ \mathbf{B}_v \boldsymbol{\tau}_v - \mathbf{A}^T \lambda - \mathbf{M}_{v2}(q_r,q_v)\ddot{q}_r - \mathbf{R}_v(q_r,q_v)\ddot{q}_r \quad (11)$$

## 3. The Proposed Controller Design

The proposed RAC-AFC controller as shown in Fig. 2 is made up of two controllers that could be theoretically designed independently. RAC was specifically designed to handle the entire kinematic problem of the MM while AFC (that was incorporated serially to the RAC) facilitates the dynamic aspect.

### 3.1 RAC

The RAC section exists in the outermost loop of the proposed controller and consists of the five output equations for $q = [x_F, y_F, \varphi, x_E, y_E]^T$:

$$\ddot{x}_{Fe} = \ddot{x}_{Fref} + K_d(\dot{x}_{Fref} - \dot{x}_{Fact}) + K_p(x_{Fref} - x_{Fact}) \quad (12)$$

$$\ddot{y}_{Fe} = \ddot{y}_{Fref} + K_d(\dot{y}_{Fref} - \dot{y}_{Fact}) + K_p(y_{Fref} - y_{Fact}) \quad (13)$$

$$\ddot{\varphi}_e = \ddot{\varphi}_{ref} + K_d(\dot{\varphi}_{ref} - \dot{\varphi}_{act}) + K_p(\varphi_{ref} - \varphi_{act}) \quad (14)$$

$$\ddot{x}_{Ee} = \ddot{x}_{Eref} + K_d(\dot{x}_{Eref} - \dot{x}_{Eact}) + K_p(x_{Eref} - x_{Eact}) \quad (15)$$

$$\ddot{y}_{Ee} = \ddot{y}_{Eref} + K_d(\dot{y}_{Eref} - \dot{y}_{Eact}) + K_p(y_{Eref} - y_{Eact}) \quad (16)$$

The subscripts *ref*, *act* and *e* refer to the input reference, actual output and error respectively. $K_p$ and $K_d$ are the proportional and derivative gains respectively. For application of the RAC only, the controller output parameters with subscript $e$ could be directly connected to the actuators input. All controller output equations in Eqs. (12) to (16) have negative feedback elements that contribute to the generation of relevant error signals that are subsequently coupled with the respective controller gains. In the global MM motion control, the controller equations can be considered as separated controls but to be executed simultaneously in real-time.

### 3.2 AFC

The AFC part constitues the inner loop of the overall control scheme. It was designed to operate in the acceleration mode of each motor at the angular side of each wheel of the platform and the joints of the arm. The AFC loop using a fixed value of the inertia matrix **IN** is shown in Fig. 3. AFC has been successfully implemented to a number of dynamical systems both theoretically as well as experimentally (Hewit and Marouf, 1996, Mailah, 1998, Kwek *et al.*, 2003) using a number of techniques to estimate the important inertial parameter necessary for the compensation of disturbances or uncertainties.

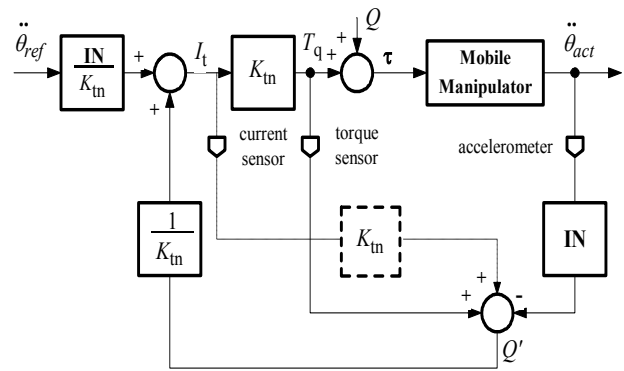

Fig. 3. A schematic of the AFC loop

Basic or crude approximation method was the simplest mode of estimation which seems to work well for most cases. Further developments in this area exploit the use of artificial intelligence (AI) technique to estimate the inertia matrix automatically and continuously using neural network (Mailah, 2001), iterative learning (Loo *et* al., 2003) and knowledge-based (Pitowarno, et al., 2002). However, the use of AI in practical application is usually not simple and often computationally intensive. It is useful to note that AFC is practically implemented through the actual measurements of the force (or current) and acceleration quantities using suitable transducers.

From Newton's second law of motion for a rotating mass, the sum of all torques ($T$) acting on the body is the product of the mass moment of inertia ($I$) and the angular acceleration ($\alpha$) of the body in the direction of the applied torque, can be represented as $\sum T = I\alpha$.

With reference to Fig. 3, the simplified dynamic model of the system can be written as

$$\tau = T_q + Q = I(\theta)\ddot{\theta}_{act} \qquad (17)$$

where $\tau$ is the total applied torque, $T_q$ is the actuated torque (motor), $Q$ is the disturbance torques, $I(\theta)$ is the mass moment of inertia of the wheels and arms (of the mobile manipulator), and $\theta$ is the angle at each wheel or joint, $\ddot{\theta}_{act}$ is the angular acceleration of the moving body.

A measurement of $Q'$ (i.e., an estimate of the disturbances, $Q$) can be obtained such that

$$Q' = T_q' - \mathbf{IN}'\ddot{\theta}_{act} \qquad (18a)$$

considering the use of a torque sensor or alternatively, if a current sensor is utilized then the equation becomes

$$Q' = I_m' K_{tn} - \mathbf{IN}'\ddot{\theta}_{act} \qquad (18b)$$

where the superscript $'$ denotes a measured or estimated quantity. The torque $T_q'$ can be measured directly using a torque sensor or indirectly by means of a current sensor (shown by dotted lines and dashed box in Fig. 3). $I_t'$ is the measured torque current and $K_{tn}$ is the motor torque constant. Parameter $\ddot{\theta}_{act}$ can be measured using an accelerometer. $\mathbf{IN}'$ may be estimated by assuming a perfect model, a fixed value through crude approximation method or other suitable techniques. It has been ascertained that if the measured or estimated values of the parameters in Eq. (18a) or (18b) were appropriately acquired, a very robust system that totally rejects the disturbances is achieved (Hewit and Burdess, 1981).

Note that in this study, a fixed value of the inertia matrix is deliberately used.

Further, by manipulating Eq. (18b) and taking into account the positive feedback element $Q'$ (estimated disturbance torque), it can be easily shown that the actual applied torque to drive the system can be expressed as

$$\tau = \mathbf{IN}'(\ddot{\theta}_{ref} - \ddot{\theta}_{act}') + I_t' K_{tn} + Q \qquad (19)$$

Eq. (19) implies a form of proportional (P) controller with regards to the use of the acceleration error signal. In this context, $\mathbf{IN}'$ can be considered as a proportional constant. It is a well known fact that the proportional controller works without extra dynamic parameter, i.e., it has insufficient capability to improve the steady state error of the system due to the system's dynamics and uncertainties (Franklin et al., 1994). The addition of an integral (I) component may improve the system performance by forcing the steady state error to minimum condition.

### 3.3 Proposed PIAFC Design

By assuming Eq. (19) is a local proportional control of the acceleration and considering that the disturbances are highly nonlinear, varied and unpredictable, a modification of the AFC scheme by incorporating an integral (I) component to the inertia matrix estimator is proposed. This is referred to as PIAFC. The integral feedback controller equation can be written as

$$G_c(t) = \left[\int_0^t e(t)dt\right]\mathbf{IN}_I \qquad (20a)$$

or in Laplace domain,

$$G_c(s) = \frac{\mathbf{IN}_I}{s} \qquad (20b)$$

where $G_c$ is the control signal and $\mathbf{IN}_I$ is an integral constant.

If the error $e(t)$ is relatively constant, $G_c(t)$ will become large and will hopefully corrects the error. By letting $e(t) = (\ddot{\theta}_{ref} - \ddot{\theta}_{act})$ and then incorporating Eq. (20a) into Eq. (19) to include the additional integral element, the proposed algorithm is given by,

$$\begin{aligned}
\tau = \mathbf{IN}_P (\ddot{\theta}_{ref} - \ddot{\theta}_{act}) + \\
\mathbf{IN}_I \int_0^t (\ddot{\theta}_{ref} - \ddot{\theta}_{act})dt + I_t' K_{tn} + Q
\end{aligned} \qquad (21)$$

where $\mathbf{IN}_P$ is a proportional constant and $\mathbf{IN}_I$ is an integral constant. Fig. 4 shows a schematic of the proposed scheme with PIAFC. The elements within the dashed box indicates the inertia matrix estimator of the AFC. Referring to Fig. 2, by substituting $\ddot{\theta}_{ref}$ in Eq. (21) with the representation of inverse kinematic output equations of Eqs. (12) to (16) the overall RAC-PIAFC output equations can be easily derived. It is obvious that Eq. (21) has the potential to minimise the computational burden.

In brief, it can be inferred that the complete control (command) equation of each actuator in the MM through the proposed controller design can be simply solved using a classic form representation as described by Eq. (21) with minimal computational burden apart from the coordinate transformation procedure required by the RAC component.

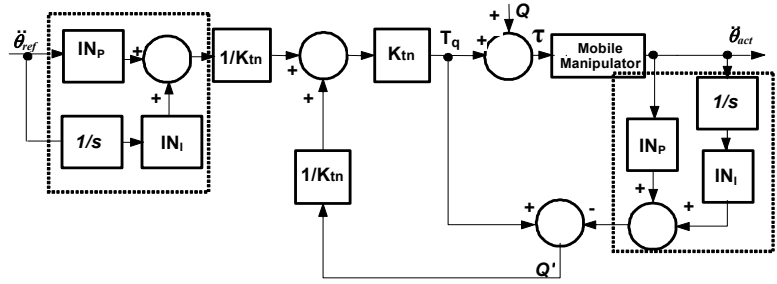

Fig. 4. The proposed PIAFC

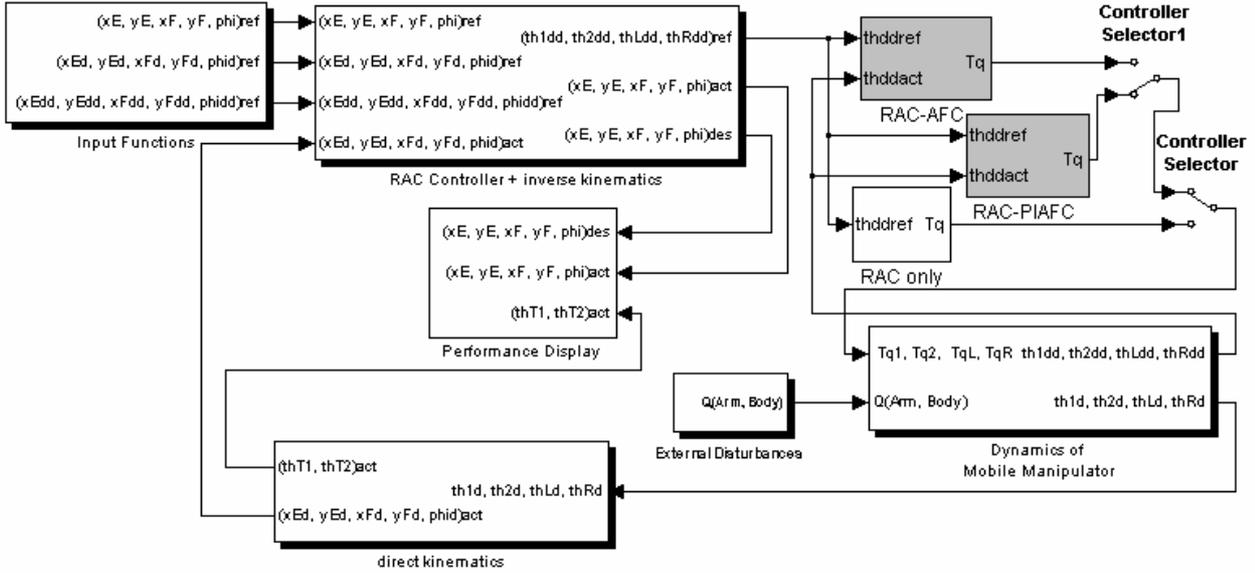

Fig. 5. A Simulink model of the scheme with three control modes - RAC, RAC-AFC and proposed RAC-PIAFC

In actual application, this has the effect of reducing the overall computational cost and time dramatically, in contrast to the use of complex AI algorithms such as those found in adaptive neural network, genetic algorithm or fuzzy logic systems which are indeed computationaly intensive and time consuming.

## 4. Simulation

The simulation was performed using MATLAB and Simulink software package. Fig. 5 shows a Simulink model of the proposed scheme that represents the input function, RAC controller, inverse and direct kinematics, dynamics of the MM, and the controller selectors. It should be noted that the actual simulation was performed in three modes, i.e., RAC only, RAC-AFC and RAC-PIAFC. The aim is to clearly demonstrate the advantages of the proposed scheme by comparison. The given task of the MM was to move its platform in a circular motion with a curvature radius of 10 m, at a speed of 0.2 m/s (at point F) and the initial heading angle orientation of $\pi/2.4$ rad to the zero angle of the world Cartesian coordinate. The manipulator was commanded to follow a specified curve track at the right-hand side of the platform starting from (10.41,0.35) of the world Cartesian coordinate. The initial tip position was set to point (10.55,0.35).

The initial experiment was conducted to determine the appropriate values of $K_p$ and $K_d$ of the RAC section. The tuning process was performed in the RAC mode using heuristic method considering some disturbances in the process. By using the tuned $K_p$ and $K_d$ values, a number of experiments was then performed to include the AFC and PIAFC schemes.

The (basic) **IN** of the RAC-AFC scheme was also approximated in the same manner. For the mobile platform, the range of the **IN** value was manually optimized from 1 to 2.8 kgm² using a step of 0.1 kgm². For the manipulator, the range of the **IN** value was from 0.01 to 0.05 kgm². Finally, by completing the tuning process for RAC and RAC-AFC, the tuning of **IN$_P$** and **IN$_I$** in the RAC-PIAFC mode can be performed.

## 5. Results and Discussion

From the initial investigation on RAC, the optimum $K_p$ and $K_d$ for $q = [x_F, y_F, \varphi, x_E, y_E]^T$ were obtained and presented in the form of diagonal matrices as follows:

$K_p = diag\{450 \ 450 \ 0.004 \ 325 \ 325\}$
$K_d = diag\{320 \ 320 \ 0.001 \ 260 \ 260\}$

It should be noted that the $K_p$ and $K_d$ values (0.004 and 0.001 respectively) for the robot's heading angle control are relatively very small compared with those of the robot's movement control. This is due to the fact that the heading angle control is very sensitive in the proposed $(x_v, y_v, \varphi)$ kinematic control mode. These values would subsequently be used in the next investigation employing the AFC scheme, initially to tune **IN** which was obtained as

$$\mathbf{IN} = diag\{0.0925 \ \ 0.0925 \ \ 2.4 \ \ 2.4\}$$

The above was then used as the reference **IN** for testing the robustness of the proposed RAC-AFC scheme. The $\mathbf{IN_P}$ and $\mathbf{IN_I}$ were then tuned by considering only a small value adjustment that was based on the tuned **IN**, and the results obtained were,

$$\mathbf{IN_P} = diag\{0.125 \ \ 0.125 \ \ 2.4 \ \ 2.4\}$$
$$\mathbf{IN_I} = diag\{0.03 \ \ 0.03 \ \ 0.01 \ \ 0.01\}$$

Apart from the condition without disturbance, there are two types of disturbances considered in the study. The first is in the form of vibratory excitation that are correspondingly applied to each wheel and joint and another the application of impact forces. Figs. 6 (a) to (c) show the models of the introduced disturbances.

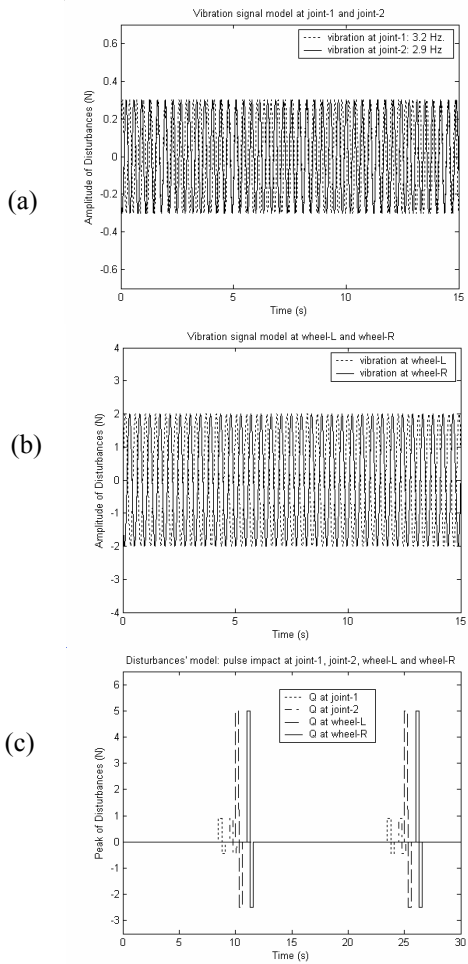

Fig. 6 (a) Vibration at the joints, (b) Vibration at the wheels and (c) Impact disturbances

Vibration was introduced into the system at a frequency of 2.2 Hz with amplitude ±2N and was applied to each wheel at different phases. For the arm, the vibration disturbances were set to 3.2 Hz, ±0.3N and while for for both joint-1 and joint-2 to 2.9 Hz, ±0.3N. These are explicitly shown in Figs. 6(a) and (b). The impact forces were introduced to joint-1, joint-2, wheel-L and wheel-R consecutively as shown in Fig. 6(c). It represents the conditions of the MM in which the arm's tip is encountering an instantaneous 'collision' with an obstacle along its path or the MM is hitting a 'bump' or 'hole' while navigating.

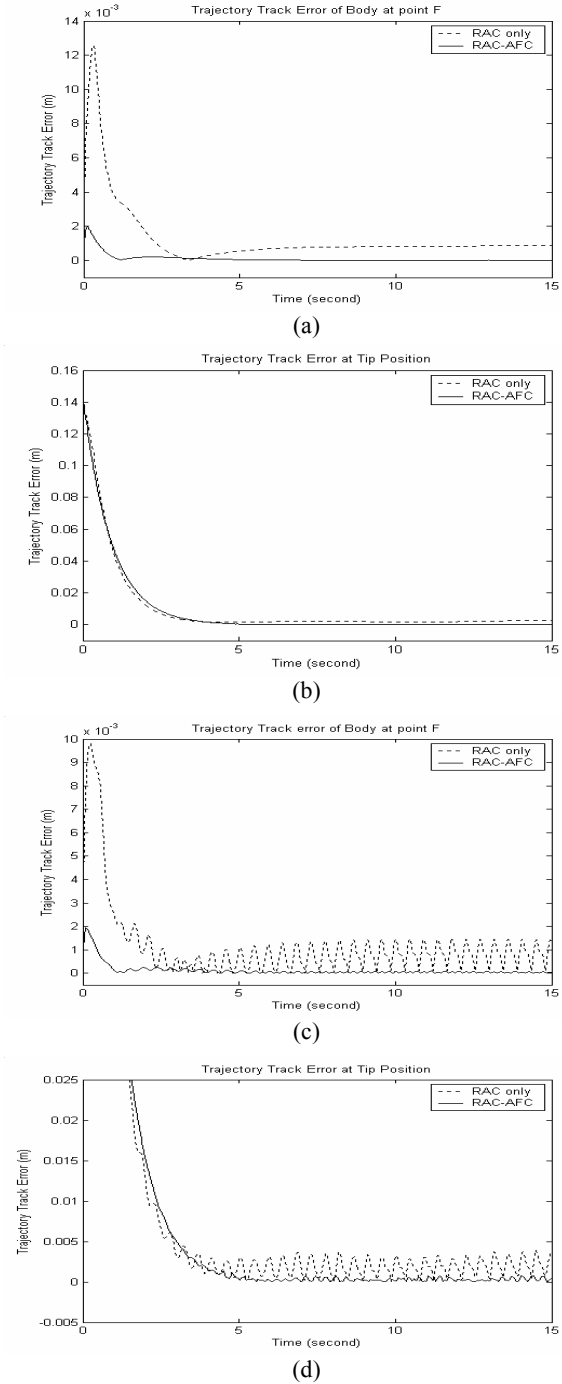

Fig. 7 (a) & (b) Track errors without disturbance (c) & (d) Track errors with vibration

The initial simulation results are shown in Figs. 7(a) to (d). The results describe the performance of RAC and RAC-AFC schemes with and without disturbances and are presented to highlight the significant improvement in applying AFC into RAC prior to discussing the advantages of the proposed RAC-PIAFC. It is very obvious that the RAC-AFC method is much more robust than the RAC method in compensating the disturbance effects. The track error generated through this scheme is far less than the RAC counterpart. The average track error for the RAC method is around 3 mm whereas for the RAC-AFC, the error is successfully suppressed to less than 1 mm mark. Having shown the effectiveness of the AFC over RAC systems, it is logical to proceed to the proposed RAC-PIAFC scheme to investigate further improvements that can be observed through the system performance. Fig. 8 shows the effect of vibration excitation to the RAC-AFC and RAC-PIAFC schemes. From the figure, it is clear that the potential of incorporating the integral term into the AFC is realized. Significant improvement on the overall performance was definitely observed. The vibration effects that occurred at peaks of around 0.5 mm for the RAC-AFC scheme has been 'rejected' to a much lower level of less than 0.2 mm in average.

Thus, the additional integral (I) element embedded into the controller through the acceleration feedback control mechanism proves its positive impact on the robust performance of the system. For a high precision robot task, the contribution of the RAC-PIAFC method is significant and indeed has real world physical implication.

Another test on the robustness of the RAC-PIAFC scheme is through a study on the effect of the introduced impact disturbances. Fig. 9 depicts the control system performances considering this condition.

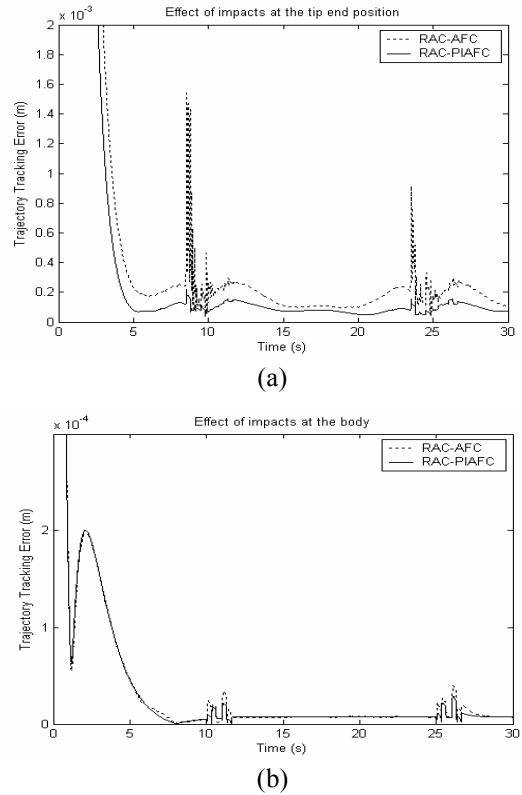

(a)

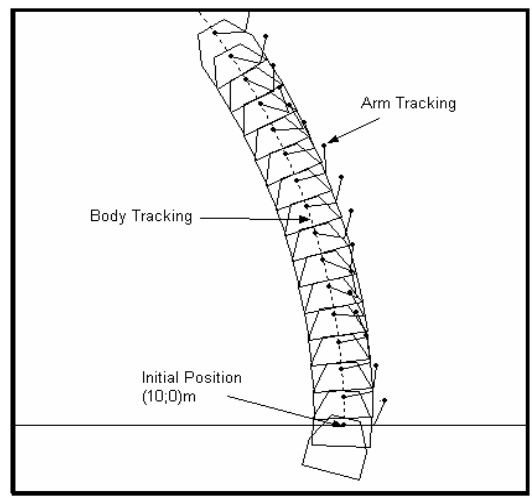

(b)

Fig. 9 (a) Effect of the impacts at the tip position and (b) Effect of impacts on the body at point F

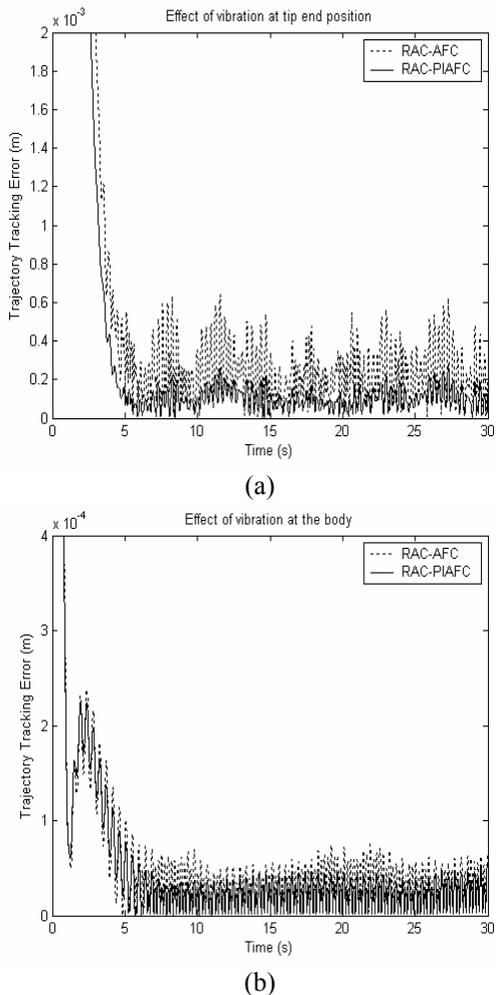

(a)

(b)

Fig. 8 (a) Effect of vibration at the tip position and (b) Effect of vibration on the body at point F

Fig. 10. MM tracking the prescribed trajectory

The impacts shown at the tip position was considerably rejected by the proposed RAC-PIAFC, as shown in Figs. 9(a) and (b). The effect was almost totally rejected by the control scheme, as indicated by the solid line in Fig. 9(a). Fig. 10 shows the path taken by MM as it executes the prescribed task.

## 6. Experimental Results

The effectiveness of the proposed RAC-PIAFC was also experimented using a a developed mobile manipulator prototype as shown in Fig. 11.

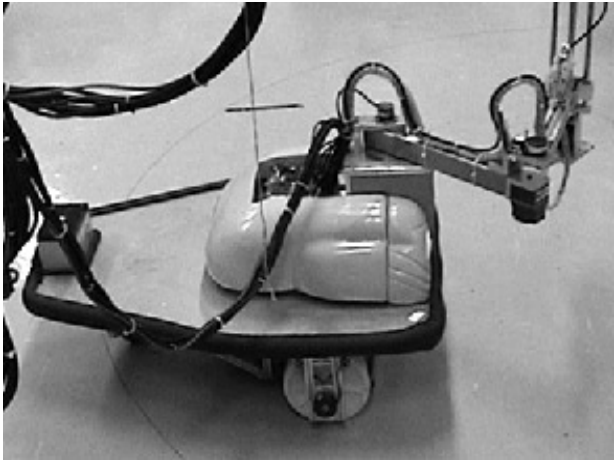

Fig. 11. A photograph of the MM prototype

The physical MM prototype shown in Fig. 11 was developed by considering most of the parameters defined in the simulation study. The rig was designed and developed using full mechatronic approach involving the integration of mechanical engineering, electrical/electronic and computer control disciplines.

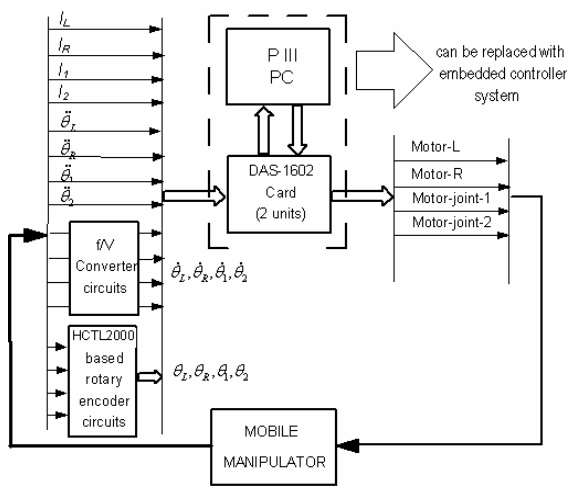

Fig. 12. A schematic of the PC-based MM control

The measurement from the sensors was used as a basis for producing the graphics of the experimental results through data acquisition procedure. Fig. 12 shows a schematic diagram of the PC-based MM control. Two data acquisition cards, DAS-1602 slotted into the motherboard of a Pentium-III PC were used to produce 12 channels of the analog input and 4 channels of analog output. The rig was also developed to operate without cable using an embedded controller by dismounting the PC-based (indicated by the dash box) and replacing it with a PIC16F877 chip. A number of computer programs written in C that were developed from the theoretical study and later performed through the simulation study were implemented and executed through experiments for both PC-based and embedded systems. Fig. 13 shows a graphic display of the Real-Time Control and Monitoring of MM's AFC On-line System (RTCM-MMAFC-OS) that was developed in this study. Through this interactive display, users can manually set the control parameter values appropriately. When the robot is in operation, i.e., executing a tracking task, a number of on-line measurements (from the sensors) can be displayed on the screen in real-time. The graph shown in the middle of the figure depicts the on-line tracking of the MM. The inner circle is the body tracking, while the outer loop is the arm tracking.

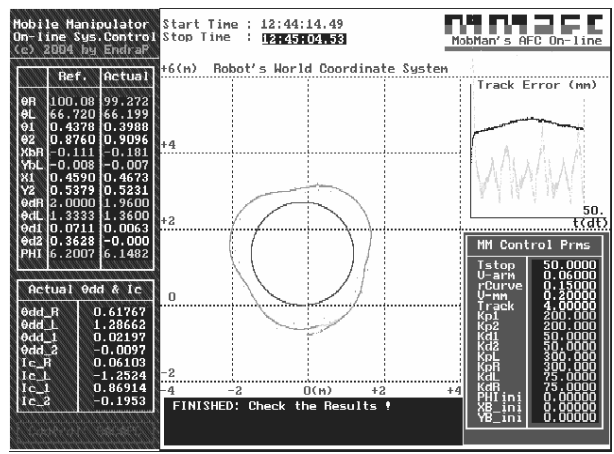

Fig. 13. A display of the RTCM-MMAFC-OS.

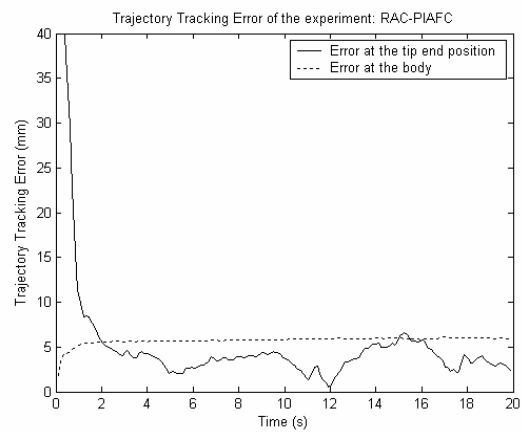

Fig. 14. Trajectory track error for the RAC-PIAFC scheme applied to physical MM

Fig. 14 shows the result of the experimental MM tracking a specified curve for 20 seconds at a body speed of 0.2 m/s. The robot task is to move in a circular path in counter-clockwise direction (just like simulation) with a radius of 1.5 (though not 10m as in simulation due to space limitation) and the arm (tip position) should follow a specified curvature trajectory at position where the robot poses the arm at the right side.

For the given robot overall dimension of $80 \times 80$ cm and the task of tracking a curve with a 1.5 m radius, the trajectory track error generated for the control schemes are not significant considering a perfectly normal mobile manipulator movement. The result shows that the track error is considerably small – around less than 5 mm (at the arm) in the experiment. It also implies that the RAC-PIAFC scheme applied to MM is feasible and practical.

## 7. Conclusion

The effectiveness of the proposed RAC-PIAFC has been demonstrated in this paper, both through simulation and experimental studies. The robustness of the proposed system was particularly highlighted in the simulation study. The RAC method is found to be readily implemented in the MM system to address the effective kinematic control, thereby establishing another alternative form of control method apart from the more popular existing ($v$, $\omega$) control. Further, combining the RAC with AFC or PIAFC to the motion control of mobile manipulator is considered a new approach in this area of study. The potentials of the RAC-PIAFC method particularly as the disturbance rejection scheme were clearly demonstrated in the study. However, further experimentation needs to be carried out to explore the maximum potentials of the scheme when other different tasks, parameters or operating and loading conditions are considered. The practical issues related to the physical MM should also be further investigated.

## Acknowledgements

We would like to thank the *Malaysian Ministry of Science, Technology and Innovation* (MOSTI), *Universiti Teknologi Malaysia* (UTM) and *Institut Teknologi Sepuluh Nopember* (ITS), Surabaya, for their continuous support in the research work. This research was fully supported by an IRPA grant (No. 03-02-06-0038EA067).